\documentclass[letterpaper, 10 pt, conference]{ieeeconf}  

\usepackage{hyperref}
\usepackage{subfig}
\usepackage{graphicx}
\usepackage{xcolor}

\IEEEoverridecommandlockouts                              

\title{\LARGE \bf
qiBullet, a Bullet-based simulator for the Pepper \\and NAO robots
}

\author{Maxime Busy$^{1}$ and Maxime Caniot$^{2}$
\thanks{$^{1}$Maxime Busy is with the Innovation Department of SoftBank Robotics Europe, France. 
        {\tt\small maxime.busy@softbankrobotics.com}}%
\thanks{$^{2}$Maxime Caniot is with the Innovation Department of SoftBank Robotics Europe, France.
        {\tt\small maxime.caniot@softbankrobotics.com}}%
}

\begin{document}

\maketitle
\thispagestyle{empty}
\pagestyle{empty}

\begin{abstract}

The Pepper and NAO robots are widely used for in-store advertizing and education, but also as robotic platforms for research purposes. Their presence in the academic field is expressed through various publications, multiple collaborative projects, and by being the standard platforms of two different RoboCup leagues. Developing, gathering data and training humanoid robots can be tedious: iteratively repeating specific tasks can present risks for the robots, and some environments can be difficult to setup. Software tools allowing to simulate complex environments and the dynamics of robots can thus alleviate that problem, allowing to perform the aforementioned processes on virtual models. One current drawback of the Pepper and NAO platforms is the lack of a physically accurate simulation tool, allowing to test scenarios involving repetitive movements and contacts with the environment on a virtual robot. In this paper, we introduce the qiBullet simulation tool, using the Bullet physics engine to provide such a solution for the Pepper and NAO robots.

\end{abstract}

\section{INTRODUCTION}

The use of the Pepper and NAO robots as research platforms is widely spread across various academic fields. Moreover, each of these robot is a standard platform in a league of the Robocup\cite{kitano1995robocup} competition (the Robocup@Home~\cite{robocupathome} league for Pepper, and the Robocup soccer~\cite{soccer_spl} for NAO). Additionally, the Pepper and NAO robot are respectively standard platforms in the World Robot Summit~\cite{wrs} competition and in the NAO Challenge~\cite{nao_challenge}. In the recent years, tremendous progresses have been achieved in the robotics field through machine learning approaches, and in particular data-driven approaches such as Deep Learning and Reinforcement Learning. Such approaches often compel the developer to gather data by iteratively repeating specific tasks with a robot. Gathering this data in a simulation able to handle complex environments and the dynamics of the simulated robot would alleviate the data gathering task, and prevent the real robotic platform from being damaged. Such a simulation would also allow to identify potential problems in a scenario including Pepper or NAO before deploying it into the real world.

\begin{figure}[htbp]
\centering
\subfloat[\label{envs_a}]{
    \includegraphics[height=3.3cm]{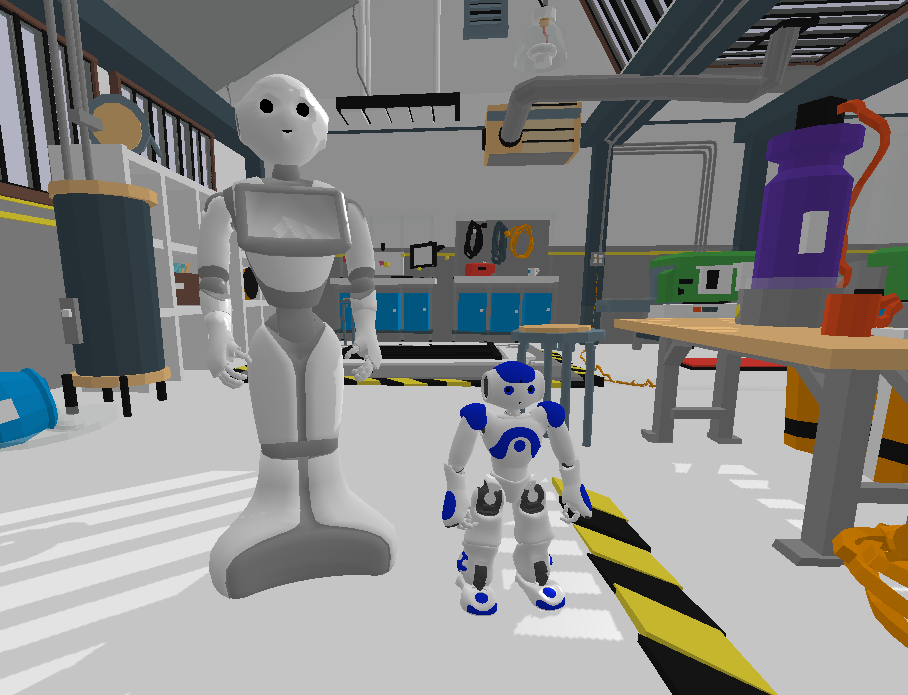}}
\hfill
\subfloat[\label{envs_b}]{
    \includegraphics[height=3.3cm]{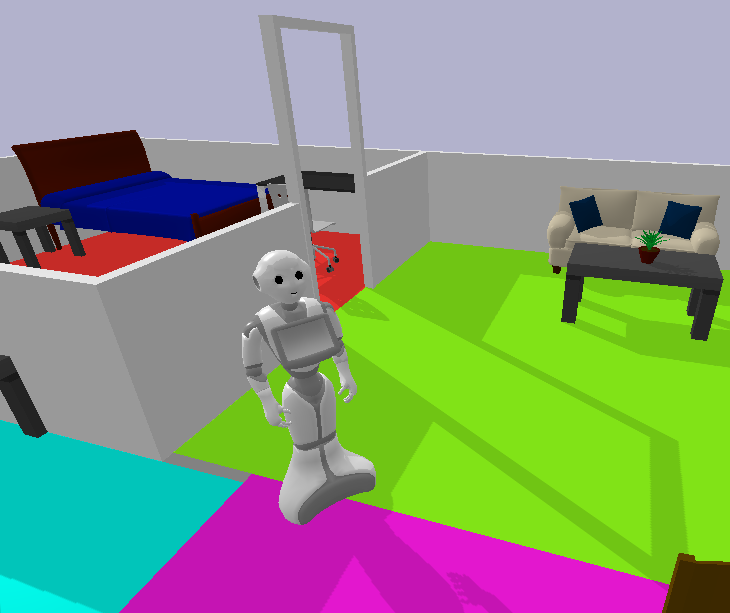}}\\
\subfloat[\label{envs_c}]{
    \includegraphics[height=3.5cm]{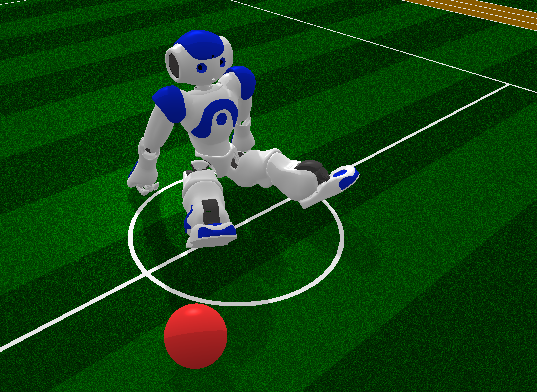}}
\hfill
\subfloat[\label{envs_d}]{
    \includegraphics[height=3.5cm]{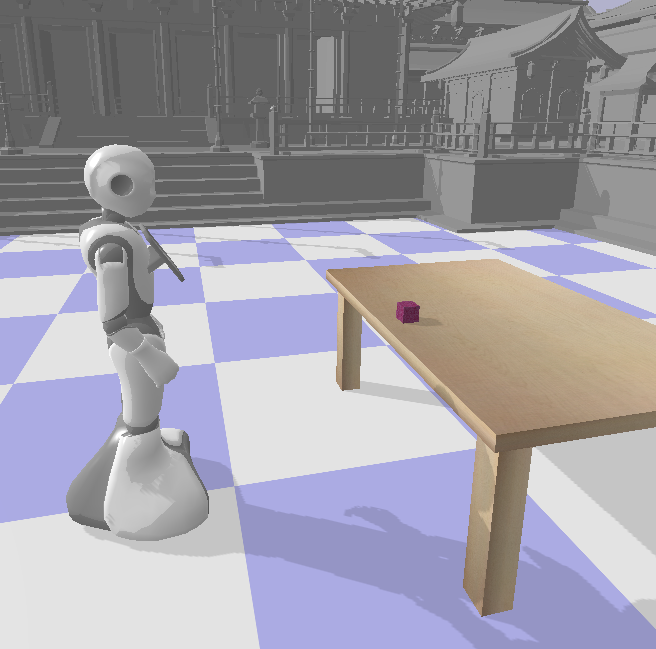}}
\caption{(a) Pepper and NAO robots in the botlab environment~\cite{botlab}. (b) Pepper robot in the Montreal arena of the Robocup@Home. (c) NAO robot in a soccer environment. (d) Pepper robot in a grasping environment.}
\label{fig:envs}
\end{figure}

Presently, Pepper and NAO models are available in simulation tools such as Gazebo~\cite{koenig2004gazebo}, V-REP~\cite{rohmer2013vrep}, Webots~\cite{michel2004webots}, or Choregraphe~\cite{pot2009choregraphe}. These implementations either lack the ability to accurately handle the physics of the model or to simulate complex environments. More recently, a Morse-based~\cite{echeverria2011morse} simulation for the Pepper robot~\cite{lier2018tobi} has been announced, but focuses on human-robot interactions and not on reliable physics.

In this paper, we introduce qiBullet, an open-source simulation tool based on the Bullet physics engine~\cite{coumans2010bullet} and the PyBullet module~\cite{coumans2019}, designed to answer the aforementioned problems. This simulator aims to provide a cross-platform and transparent mean to embed a virtual Pepper or NAO robot in different evironments (Figure~\ref{fig:envs}), via a Python-based API mimicking NAOqi~\cite{naoqi} or a ROS interface emulating the naoqi\_driver\footnote{\url{http://wiki.ros.org/naoqi_driver}} ROS~\cite{ROS} package. In order to describe our work, we first specify the strong ties between qiBullet and the Bullet physics engine. We then describe the simulator itself, its interfaces, the components of a virtual robot model and its control and sensing capabilities, along with scenarios showcasing the tool. Lastly, we discuss the exploitation and availability of the simulator.

\section{PHYSICS ENGINE}
We based our approach on a comparison of physics-based simulation libraries proposed by Erez \textit{et al.}~\cite{erez2015simulationtools}, discussing the differences between the Bullet, Havok~\cite{Havok2018}, MuJoCo\cite{todorov2012mujoco}, ODE\cite{ODE2019Jan} and Physx\cite{PhysX2018Nov} physics engine.

To build our simulator, we chose the Bullet physics engine and the additional PyBullet module. The physics engine is integrated with many of the popular robotics software platforms, such as V-REP and Gazebo, and presents the advantage of being open-source. This engine can additionally be extended with PyBullet, an open-source Python module providing robotics and machine learning capabilities~\cite{tan2018simtoreal}~\cite{breyer2019comparing}~\cite{choromanski2019random}~\cite{dalin2019learning}~\cite{zeng2019tossingbot}.

The qiBullet simulation has been designed to inherit the cross-platform properties of the PyBullet Python module and Bullet physics engine: the simulation tool can be run on Linux, Windows and MacOS.

\section{SIMULATOR}
In this section, we will detail how a virtual robot is defined in the qiBullet simulator.

\subsection{Robot model}
We use a Unified Robot Description Format (URDF)~\cite{urdf} file to describe the model of a virtual robot. The different links of the model, their masses, inertia matrices and the joints connecting them are extracted from this file by the engine. Mesh files are associated to each link, allowing the engine to render the visual aspect of the robot model and to perform collision checking.

\subsection{Interface}
Two different ways of interacting with the robot virtual model are proposed in the simulation: The Python-based qiBullet API or the ROS Framework. \\

The Python-based qiBullet API, built on top of the Pybullet API, allows the user to interact with the simulated robot and more generally with the simulated environment. To ease the integration into existing projects and ensure code consistency, the qiBullet API mimics a part of the NAOqi API, rendering the interaction with a virtual and a real robot as transparent as possible.

\begin{figure}[htbp]
\centerline{\includegraphics[scale=0.3]{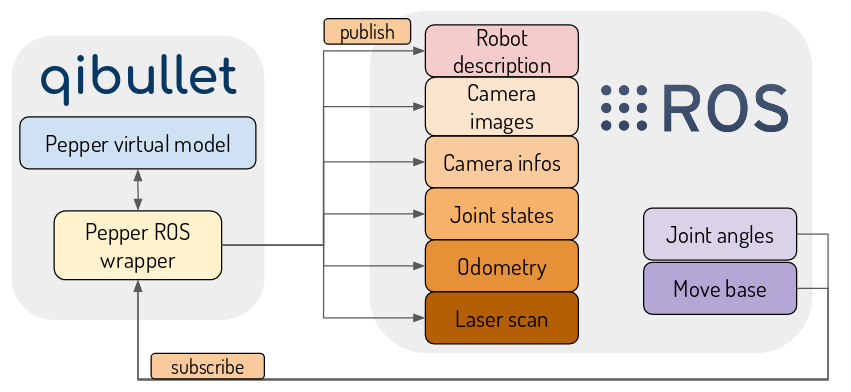}}
\caption{Interfacing of a Pepper virtual model with the ROS framework. The wrapper specifies the description of the robot, publishes the sensory data of the model and subscribes to control topics.}
\label{fig:ros_wrapper}
\end{figure}

The qiBullet simulator also provides a "ROS wrapper" built on top of the API, emulating the naoqi\_driver ROS package and allowing the user to interact with the virtual robot model through the ROS framework. As described in \autoref{fig:ros_wrapper}, the wrapper retrieves the sensory data of the model to publish it into the framework, and retrieves commands from the framework to apply them onto the model.

In order to facilitate the use of the simulator for machine learning tasks, the API can also be used to instantiate, reset, and stop several independent simulated instances, running in parallel. It can also be used to spawn or remove virtual robots from an instance, and to control the position of the light source of the simulation.

\subsection{Control and Sensing}
This subsection illustrates the control and sensing capabilities of the Pepper and NAO virtual robots in the qiBullet simulator.\\

The joints of the models can be controlled independently or as groups to reach a specified articular position with a specified speed. Several postures can be applied onto the models, similar to the postures defined within the NAOqi API. Additionally, the base of the Pepper virtual robot can be controlled in position or in velocity.

\begin{figure}[htbp]
\centerline{\includegraphics[scale=0.18]{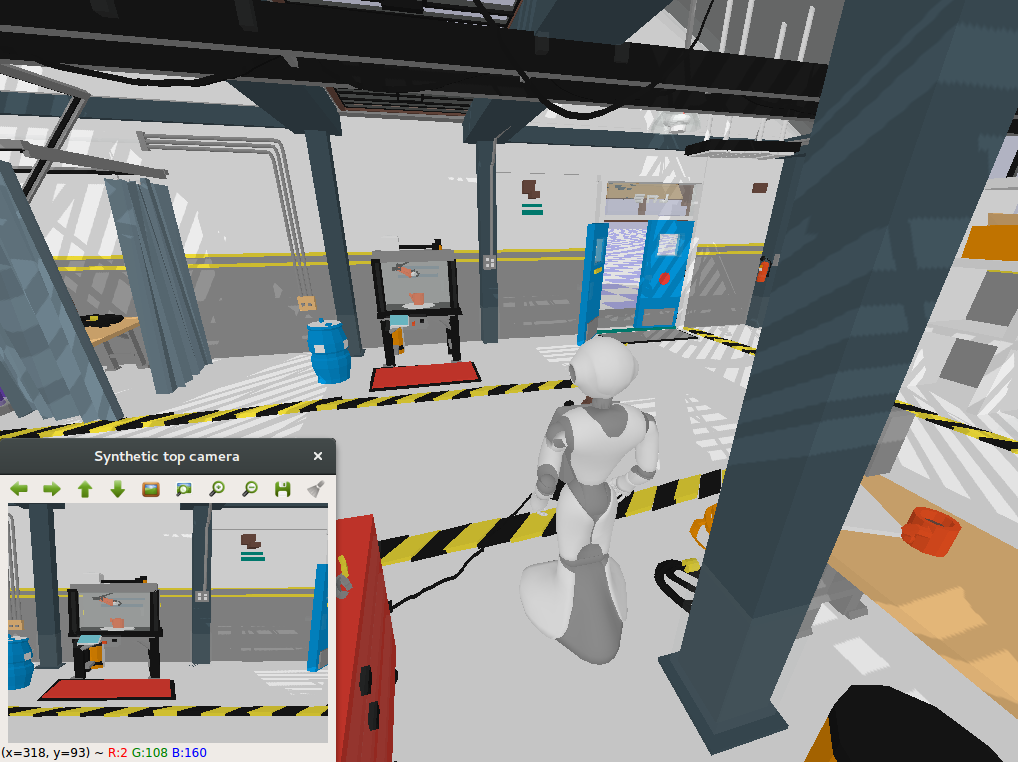}}
\caption{Synthetic RGB image retrieval from the top camera of a virtual Pepper, in a simulated environment. The obtained RGB image is displayed in the bottom left corner, using OpenCV.}
\label{fig:rgb_camera}
\end{figure}

Each Pepper and NAO virtual model embeds two RGB cameras (\autoref{fig:rgb_camera}). The Pepper virtual model additionally embeds a depth camera. Similarly to the NAOqi API, the resolution of the retrieved synthetic images can be selected by the user. The parameters of the simulated cameras are tuned to match the ones of the real model, although it is important to point out that a real depth image from Pepper will be more noisy than a synthetic one. The Pepper virtual model also possesses laser sensors attached to its base, mimicking the lasers of the real robot. Finally, the position of each model in the world frame can be directly retrieved from the qiBullet API, providing odometry information. In our simulation, the odometry drift is not simulated. The API can also provide collision data for a link or a group of links of the desired virtual model.

\subsection{Scenarios}
In this subsection, we present three different scenarios showcasing the qiBullet simulator. \\

\begin{figure}[htbp]
\centering
\subfloat[\label{grasp_a}]{
    \includegraphics[height=4.4cm]{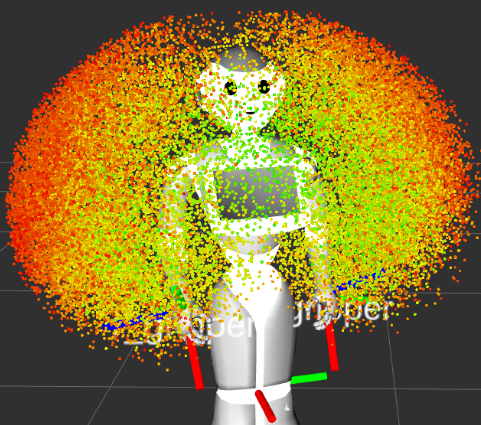}}
\hfill
\subfloat[\label{grasp_b}]{
    \includegraphics[height=4.4cm]{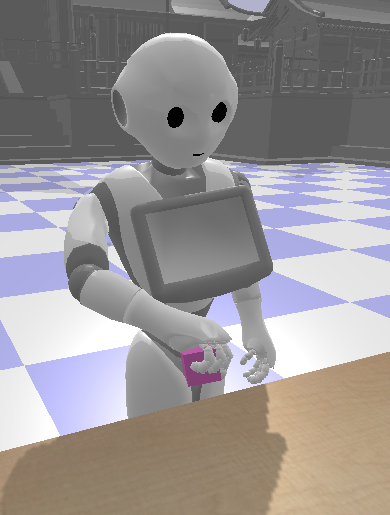}}
    
\caption{(a) RViz display of the right and left arm workspaces of the Pepper robot, with kinematic chains starting from the Tibia link, and respectively ending at the right and left gripper links. The color of each point represents the associated normalized manipulability value: green corresponds to the maximum manipulability value, while red corresponds to the minimum. (b) Grasping scenario with a virtual Pepper robot: the pink cube in the right hand of Pepper is the object to be grasped.}
\label{fig:workspace_and_grasp}
\end{figure}

\paragraph{Workspace computation} We aim to sample the right and left arm workspaces of the Pepper robot, and to evaluate the manipulability~\cite{yoshikawa1985manipulability} of each sampled configuration. The kinematic chains start at the Tibia link and respectively end at the right and left gripper links. To do so, we instantiate 10 simulation instances. In each simulation and for each iteration, an articular position is randomly defined and applied onto the joints of a chain. If the chain is self colliding with the model, the position is deemed incorrect, and another position is computed. If the end effector does not self collide, the reached position and the corresponding manipulability are added to the workspace. When all of the instances have reached 4000 successful iterations for each arm, the results are combined and normalized with respect to the maximum manipulability value obtained to generate a workspace containing 40000 elements for each kinematic chain (see \autoref{fig:workspace_and_grasp}(a)).

\paragraph{Grasping task} We define a grasping scenario, where Pepper is placed in front of a table on which an object to be grasped is positioned. The physical properties of the Bullet physics engine allow to virtually test out such a scenario with the Pepper virtual model (see \autoref{fig:workspace_and_grasp}(b)). \\

\begin{figure}[htbp]
\centering
\subfloat[\label{walk_a}]{
    \includegraphics[height=3.4cm]{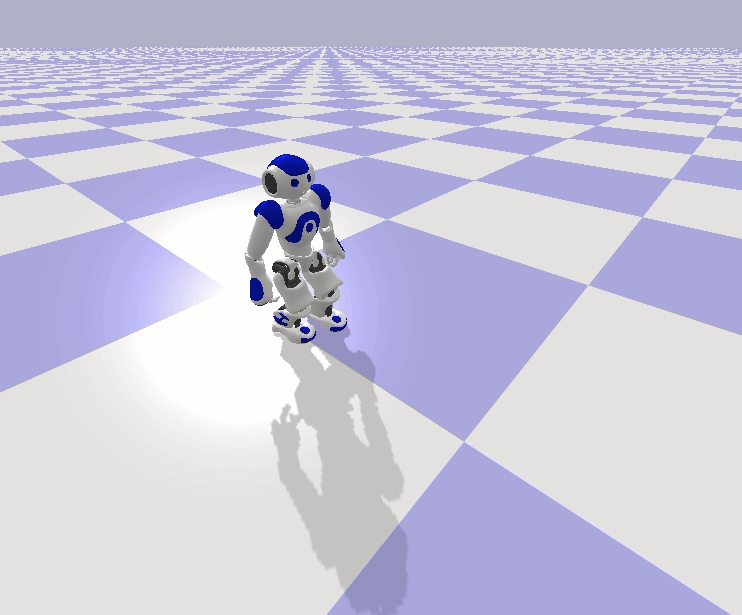}}
\hfill
\subfloat[\label{walk_b}]{
    \includegraphics[height=3.4cm]{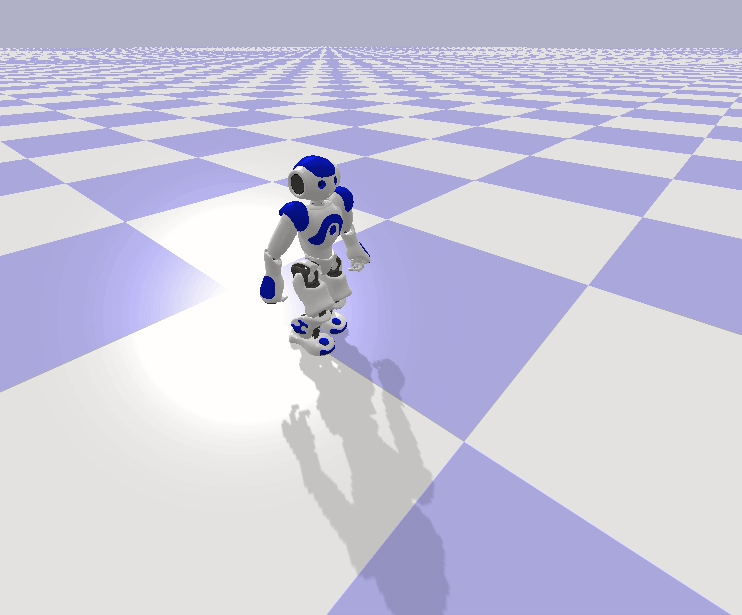}}
    
\caption{(a) and (b) Virtual NAO robot being controlled by a walking algorithm.}
\label{fig:nao_walk}
\end{figure}

\paragraph{Walking task} We define a walking scenario, where NAO is placed standing still in a flat world. The simulator can be used to test the balance of the robot while being controlled with different walking algorithms (see \autoref{fig:nao_walk}). \\

The use of the qiBullet simulator can be extended beyond these scenarios, for instance to generate/extend datasets with synthetic data~\cite{laflaquiere2019self}, to train Reinforcment Learning algorithms or to perform localization and navigation tasks.

\section{EXPLOITATION}
In order to foster the use of the qiBullet simulator, its code is made open-source and is available on Github\footnote{\url{https://github.com/softbankrobotics-research/qibullet}}. The qiBullet repository contains the files defining the Pepper and NAO robot models, the Python-based qiBullet API, examples illustrating how to use the simulator, an automatically generated documentation, and unit tests. The unit tests are automatically launched by a continuous integration tool\footnote{\url{https://travis-ci.org/softbankrobotics-research/qibullet}} when the simulation is updated, in order to ensure the stability of the library. Moreover, the qiBullet\footnote{\url{https://pypi.org/project/qibullet/}} Python package has been created and is updated after each new release of the simulator: this particular packaging allows a simple installation of our simulator and of its dependencies.

\section{CONCLUSION}
In this paper we introduce qiBullet, a simulator based on PyBullet and the Bullet physics engine, aiming to virtually emulate SoftBank Robotics' robots in a physically accurate fashion. The simulator can sustain multiple instances running in parallel, provides a Python API to interact with the sensors and actuators of the simulated models, and can be interfaced with the ROS framework. In an effort to open the simulator to the community, its code has been made public and is hosted on Github. The work on the qiBullet simulator is still ongoing, we envision introducing additional features and enhancing the existing robot models.

\bibliographystyle{unsrt}
\bibliography{biblio}

\begin{thebibliography}{10}

\bibitem{kitano1995robocup}
Hiroaki Kitano, Minoru Asada, Yasuo Kuniyoshi, Itsuki Noda, and Eiichi Osawa.
\newblock Robocup: The robot world cup initiative.
\newblock 1995.

\bibitem{robocupathome}
Thomas Wisspeintner, Tijn Van Der~Zant, Luca Iocchi, and Stefan Schiffer.
\newblock Robocup@ home: Scientific competition and benchmarking for domestic
  service robots.
\newblock {\em Interaction Studies}, 10(3):392--426, 2009.

\bibitem{soccer_spl}
Eric Chown and Michail~G. Lagoudakis.
\newblock The standard platform league.
\newblock In {\em RoboCup 2014: Robot World Cup {XVIII} [papers from the 18th
  Annual RoboCup International Symposium, Jo{\~{a}}o Pessoa, Brazil, July 15},
  pages 636--648, 2014.

\bibitem{wrs}
{World Robot Summit}.
\newblock [Online], Available:
  \textcolor{blue}{\url{https://worldrobotsummit.org/en/}}.

\bibitem{nao_challenge}
{NAO Challenge}.
\newblock [Online], Available:
  \textcolor{blue}{\url{https://www.naochallenge.it/}}.

\bibitem{botlab}
Alan Zimmerman.
\newblock Botlab environment.
\newblock [Online], Available:
  \textcolor{blue}{\url{https://poly.google.com/view/2hWxc7Hk9CJ}}.

\bibitem{koenig2004gazebo}
Nathan Koenig and Andrew Howard.
\newblock Design and use paradigms for gazebo, an open-source multi-robot
  simulator.
\newblock In {\em 2004 IEEE/RSJ International Conference on Intelligent Robots
  and Systems (IROS)(IEEE Cat. No. 04CH37566)}, volume~3, pages 2149--2154.
  IEEE, 2004.

\bibitem{rohmer2013vrep}
Eric Rohmer, Surya~PN Singh, and Marc Freese.
\newblock V-rep: A versatile and scalable robot simulation framework.
\newblock In {\em 2013 IEEE/RSJ International Conference on Intelligent Robots
  and Systems}, pages 1321--1326. IEEE, 2013.

\bibitem{michel2004webots}
Olivier Michel.
\newblock Cyberbotics ltd. webots™: professional mobile robot simulation.
\newblock {\em International Journal of Advanced Robotic Systems}, 1(1):5,
  2004.

\bibitem{pot2009choregraphe}
Emmanuel Pot, J{\'e}r{\^o}me Monceaux, Rodolphe Gelin, and Bruno Maisonnier.
\newblock Choregraphe: a graphical tool for humanoid robot programming.
\newblock In {\em RO-MAN 2009-The 18th IEEE International Symposium on Robot
  and Human Interactive Communication}, pages 46--51. IEEE, 2009.

\bibitem{echeverria2011morse}
Gilberto Echeverria, Nicolas Lassabe, Arnaud Degroote, and S{\'e}verin
  Lemaignan.
\newblock Modular open robots simulation engine: Morse.
\newblock In {\em 2011 IEEE International Conference on Robotics and
  Automation}, pages 46--51. Citeseer, 2011.

\bibitem{lier2018tobi}
Florian Lier and Sven Wachsmuth.
\newblock Towards an open simulation environment for the pepper robot.
\newblock In {\em Companion of the 2018 ACM/IEEE International Conference on
  Human-Robot Interaction}, pages 175--176. ACM, 2018.

\bibitem{coumans2010bullet}
Erwin Coumans.
\newblock Bullet physics engine.
\newblock {\em Open Source Software:
  \textcolor{blue}{http://bulletphysics.org}}, 1(3), 2010.

\bibitem{coumans2019}
Erwin Coumans and Yunfei Bai.
\newblock Pybullet, a python module for physics simulation for games, robotics
  and machine learning.
\newblock \textcolor{blue}{\url{http://pybullet.org}}, 2016--2019.

\bibitem{naoqi}
Aldebaran Robotics.
\newblock Naoqi framework.
\newblock [Online], Available:
  \textcolor{blue}{\url{http://doc.aldebaran.com/2-5/index\_dev\_guide.html}}.

\bibitem{ROS}
Morgan Quigley, Ken Conley, Brian Gerkey, Josh Faust, Tully Foote, Jeremy
  Leibs, Rob Wheeler, and Andrew~Y Ng.
\newblock Ros: an open-source robot operating system.
\newblock In {\em ICRA workshop on open source software}, volume~3, page~5.
  Kobe, Japan, 2009.

\bibitem{erez2015simulationtools}
Tom Erez, Yuval Tassa, and Emanuel Todorov.
\newblock Simulation tools for model-based robotics: Comparison of bullet,
  havok, mujoco, ode and physx.
\newblock In {\em 2015 IEEE international conference on robotics and automation
  (ICRA)}, pages 4397--4404. IEEE, 2015.

\bibitem{Havok2018}
{Havok physics engine}.
\newblock [Online], Available: \textcolor{blue}{\url{www.havok.com}}.

\bibitem{todorov2012mujoco}
Emanuel Todorov, Tom Erez, and Yuval Tassa.
\newblock Mujoco: A physics engine for model-based control.
\newblock In {\em 2012 IEEE/RSJ International Conference on Intelligent Robots
  and Systems}, pages 5026--5033. IEEE, 2012.
\newblock [Online], Available: \textcolor{blue}{\url{www.mujoco.org}}.

\bibitem{ODE2019Jan}
{Open Dynamics Engine}.
\newblock [Online], Available: \textcolor{blue}{\url{http://ode.org}}.

\bibitem{PhysX2018Nov}
{PhysX physics engine}.
\newblock [Online], Available:
  \textcolor{blue}{\url{www.geforce.com/hardware/technology/physx}}.

\bibitem{tan2018simtoreal}
Jie Tan, Tingnan Zhang, Erwin Coumans, Atil Iscen, Yunfei Bai, Danijar Hafner,
  Steven Bohez, and Vincent Vanhoucke.
\newblock Sim-to-real: Learning agile locomotion for quadruped robots.
\newblock {\em arXiv preprint arXiv:1804.10332}, 2018.

\bibitem{breyer2019comparing}
Michel Breyer, Fadri Furrer, Tonci Novkovic, Roland Siegwart, and Juan Nieto.
\newblock Comparing task simplifications to learn closed-loop object picking
  using deep reinforcement learning.
\newblock {\em IEEE Robotics and Automation Letters}, 4(2):1549--1556, 2019.

\bibitem{choromanski2019random}
Krzysztof Choromanski, Aldo Pacchiano, Jack Parker-Holder, Jasmine Hsu, Atil
  Iscen, Deepali Jain, and Vikas Sindhwani.
\newblock When random search is not enough: Sample-efficient and noise-robust
  blackbox optimization of rl policies.
\newblock {\em arXiv preprint arXiv:1903.02993}, 2019.

\bibitem{dalin2019learning}
Elo{\"\i}se Dalin, Pierre Desreumaux, and Jean-Baptiste Mouret.
\newblock Learning and adapting quadruped gaits with the" intelligent trial \&
  error" algorithm.
\newblock 2019.

\bibitem{zeng2019tossingbot}
Andy Zeng, Shuran Song, Johnny Lee, Alberto Rodriguez, and Thomas Funkhouser.
\newblock Tossingbot: Learning to throw arbitrary objects with residual
  physics.
\newblock {\em arXiv preprint arXiv:1903.11239}, 2019.

\bibitem{urdf}
Willow Garage.
\newblock Universal robot description format (urdf).
\newblock {\em Http://Www. ros. org/urdf/, 2009}, 2009.

\bibitem{yoshikawa1985manipulability}
Tsuneo Yoshikawa.
\newblock Manipulability of robotic mechanisms.
\newblock {\em The international journal of Robotics Research}, 4(2):3--9,
  1985.

\bibitem{laflaquiere2019self}
Alban Laflaqui{\`e}re and Verena~V Hafner.
\newblock Self-supervised body image acquisition using a deep neural network
  for sensorimotor prediction.
\newblock {\em arXiv preprint arXiv:1906.00825}, 2019.

\end{thebibliography}

\end{document}